\documentclass[twocolumn]{article}

\usepackage[margin=0.85in]{geometry} % Standard margins for arXiv
\usepackage{microtype}
\usepackage{graphicx}
\usepackage{subfigure}
\usepackage{booktabs}
\usepackage{hyperref}
\usepackage{url}
\usepackage{multirow}
\usepackage{times} % Use Times New Roman font (matches BERT/ACL style)
\usepackage{algorithm}
\usepackage{algorithmic}
\usepackage{tikz}
\usepackage{pgfplots}
\usepackage{xcolor}
\pgfplotsset{compat=1.18}

\usepackage[T1]{fontenc}
\usepackage[utf8]{inputenc}
\usepackage{textcomp}

\usepackage{amsmath}
\usepackage{amssymb}
\usepackage{mathtools}
\usepackage{amsthm}

\usepackage{natbib}
\setcitestyle{authoryear,round,citesep={;},aysep={,},yysep={,}}

\title{\bfseries\Large SemanticALLI: Caching Reasoning, Not Just Responses, in Agentic Systems}

\author{
  \textbf{Varun Chillara}\textsuperscript{*} \quad
  \textbf{Dylan Kline}\textsuperscript{*} \quad
  \textbf{Christopher Alvares}\textsuperscript{$\dagger$} \quad
  \textbf{Evan Wooten}\textsuperscript{$\dagger$} \quad
  \textbf{Huan Yang}\textsuperscript{$\dagger$} \\
  \textbf{Shlok Khetan} \quad
  \textbf{Cade Bauer} \quad
  \textbf{Tr\'e Guillory} \quad
  \textbf{Tanishka Shah} \\
  \textbf{Yashodhara Dhariwal} \quad
  \textbf{Volodymyr Pavlov} \\
  \textbf{George Popstefanov}\textsuperscript{$\ddagger$} \\[0.6em]
  PMG, Dallas, TX, USA \\[0.4em]
}

\date{} % Remove date

\begin{document}

\maketitle

% --- MANUAL FOOTNOTES FOR SYMBOLS ---
{
  \renewcommand{\thefootnote}{\fnsymbol{footnote}}
  \footnotetext[1]{Equal contribution.}
  \footnotetext[2]{Advisor.}
  \footnotetext[3]{Project Funder.}
}
\begin{abstract}
Agentic AI pipelines suffer from a hidden inefficiency: they frequently reconstruct identical intermediate logic, such as metric normalization or chart scaffolding, even when the user's natural language phrasing is entirely novel. Conventional boundary caching fails to capture this efficiency because it treats inference as a monolithic black box.

We introduce \textbf{SemanticALLI} (part of Alli, PMG's flagship marketing intelligence platform), a pipeline-aware architecture designed to operationalize the \emph{redundancy of reasoning}. By decomposing generation into \textit{Analytic Intent Resolution (AIR)} and \textit{Visualization Synthesis (VS)}, \textbf{SemanticALLI} promotes structured intermediate representations (IRs) to first-class cacheable artifacts. 

The impact of caching \emph{within} the agentic loop is substantial. In our evaluation, baseline monolithic caching caps at a 38.7\% hit rate due to linguistic variance. In contrast, our structured approach allows for an additional stage, the Visualization Synthesis stage to achieve an 83.10\% hit rate, bypassing 4,023 LLM calls with a median latency of just 2.66 ms. This internal reuse reduces total token consumption, offering a practical lesson for AI system design: even when users rarely repeat themselves, the pipeline often does—at stable, structured checkpoints where caching is most reliable.
\end{abstract}

\section{Introduction}\label{sec:intro}

A useful stress test for today’s enterprise LLM stacks is painfully simple: ask for a dashboard in natural language, then watch the clock. What looks like a single request in the UI typically expands into a small pipeline—schema inspection, intent resolution, metric and filter selection, and finally the synthesis of chart or dashboard code. When everything goes well, the output is impressive. When it takes a minute (or three), the impression does not last for modern enterprise users of Business Intelligence \textbf{(BI)}.

We refer to this failure mode as the \textit{Latency--Utility Gap}: the system is capable of high-utility analytic reasoning, but its latency profile pushes it outside the window in which a human would be willing to wait. In products where retention is mediated by attention rather than contractual obligation, that gap is not a rounding error; it is the adoption constraint. Work on the \textit{attention economy} makes this point directly: engagement behaves less like a linear function of quality and more like a brittle threshold phenomenon once responsiveness slips past user expectations~\cite{Roffarello2025DigitalAttention}. Empirically, interactive systems begin to hemorrhage users when response times exceed roughly 7--10 seconds~\cite{Arapakis_2021_latency_mobile,nielsen1993usability}. Many production-grade analytic agents operate far beyond that bound.

The default engineering reaction is to make the model faster. Quantization, distillation, and smaller frontier alternatives can cut wall-clock time in controlled settings, sometimes sharply. Yet this strategy has an awkward second-order cost: lighter models often fail in precisely the places enterprise analytics cannot afford to fail—grounding in data, constraint satisfaction, and compositional reasoning. Teams then compensate with prompt tuning, regression testing, and retries, and the system’s “savings”—token usage and latency—leak away in a slow drip of operational friction.

Looking more closely, the bottleneck is frequently not inference speed per se, but the \textit{redundancy of reasoning}. A meaningful fraction of enterprise questions are paraphrases of earlier intents~\cite{dunlap2025freshness,palmini2025exploringlanguagepatternsprompts}, and both latency and API cost scale with token throughput~\cite{ghaffari2025ensembleembeddingapproachimproving}. That observation motivates monolithic caching, where the cache key is the user prompt (or its embedding neighborhood) and the value is the final system output. Monolithic prompt$\rightarrow$output caching is helpful when repetition is at the surface level. But it is also brittle: once a user asks a genuinely new question, the cache either “hits” on superficial semantic similarity and returns something that is close in vector space but wrong in business semantics—or misses the cache entirely.

The more stable reuse opportunity sits inside the workflow. Analytic pipelines contain intermediate artifacts that recur even when the prompts do not: metric selections, standard dimensional groupings, and reusable artifacts. \textbf{SemanticALLI} is built around this premise. Rather than treating the agentic system as a black box, we make the internal structure explicit and cache at logical checkpoints:
\begin{enumerate}
    \item \textbf{Analytic Intent Resolution (AIR)} --- a normalized, structured representation of \emph{what} to compute (metrics, dimensions, filters, and analytic framing), abstracted away from rendering details; and
    \item \textbf{Visualization Synthesis (VS)} --- the implementation layer describing \emph{how} to render that intent (e.g., reusable chart specifications or dashboard code).
\end{enumerate}

A small example captures the idea. A user might request: ``Build an executive summary dashboard with top KPIs across media, such as sales, impressions, clicks, and spend.'' Another might type: ``Show media KPIs.'' Monolithic caching may treat these two strings as similar; however, if they fail to meet a similarity threshold $\tau$, they will ultimately miss the cache, causing complete LLM generation. \textbf{SemanticALLI} instead asks a different question: do both prompts collapse to the same intent and visualization structure (or at least share substructures)? If they do, we should not pay for that reasoning twice.

This framing pushes caching from a passive perimeter optimization into an active component of the reasoning system. It also forces a more careful retrieval design: for example enterprise analytics is full of near-misses where ``CPM'' and ``CPC'' are semantically adjacent but operationally distinct. \textbf{SemanticALLI} therefore combines exact matching with hybrid semantic/lexical retrieval, ensuring reuse is fast \emph{and} appropriately constrained by entity-level evidence.

Business Intelligence is just one example of how \textit{Structured Intermediate Representation Caching (SIRC)} can be applied to optimize Agentic AI systems. This idea can be applied more generally to any Agentic workflow in which reuse is internal agents.

\subsection{Contributions}

We make three contributions:
\begin{enumerate}
    \item \textbf{Pipeline-aware caching for agentic systems} We formalize an abstract efficiency-first systems architecture through a production example consisting of two-stage decomposition (AIR and VS) and treat intermediate reasoning artifacts as cacheable units rather than incidental byproducts.
    \item \textbf{A hybrid retrieval mechanism for high-cardinality domains.} We integrate exact hashing with dense retrieval and BM25 lexical constraints, using rank fusion and reranking (RRF) ~\cite{cormack2009rrf} to reduce semantic collisions in business terminology.
    \item \textbf{Empirical evidence that shows internal reuse usability significance.} Across controlled evaluations, we show that when prompt$\rightarrow$output reuse becomes scarce under strict similarity thresholds, substantial reuse can still be recovered at structured intermediate checkpoints—yielding meaningful reductions in LLM calls, tokens, and latency.
\end{enumerate}

Collectively, these results narrow the \textit{Latency--Utility} Gap by exploiting what the pipeline already knows: many “new” user requests share a common internal structure.

\section{Related Works}
\label{sec:related_works}

Caching has re-emerged as a pragmatic answer to the rising marginal cost of LLM-based systems: if the exact computation is likely to be requested again, store it once and reuse it. In the LLM setting, this idea appears in several guises—model-side optimizations (e.g., reuse of internal states via KVCaching)~\cite{kwon2023pagedattention} as well as application-side memoization of inputs and outputs of a model. Our emphasis is on the latter, but with a twist: \emph{where} the cache is allowed to attach to the workflow matters at least as much as \emph{how} retrieval is performed.

\subsection{Semantic Caching}
\label{sec:semanticcaching}

Semantic caching replaces exact string matching with similarity in an embedding space, making it viable to reuse answers even when users paraphrase. The enabling ingredient is a sentence-level representation model that places meaningfully related queries near each other~\cite{reimers2019sentencebertsentenceembeddingsusing}. Systems such as GPTCache~\cite{bang-2023-gptcache}, InstCache~\cite{zou2025instcachepredictivecachellm}, and GenCache~\cite{gencache_cite} operationalized this idea for LLM applications, demonstrating that nearest-neighbor retrieval over embedded prompts can reduce end-to-end latency when query repetition is common. More recent work has pushed beyond “pure similarity search” by exploiting the empirical skew of real workloads: approaches like SCALM~\cite{li2024scalmsemanticcachingautomated}, Cache Saver~\cite{cachesaver_cite}, and Asteria~\cite{ruan2025asteriasemanticawarecrossregioncaching} detect and prioritize recurring patterns to increase effective hit rates in practice.

And yet, there is a structural limitation that is easy to miss if one only benchmarks on paraphrases. A monolithic prompt$\rightarrow$output cache inherits the ambiguity of the prompt. Two prompts may be close in vector space while still differing in a single business-critical entity (``CPC'' vs.\ ``CPM'' is the canonical trap), and the retrieval layer has no native notion of which terms are non-negotiable. Put differently, semantic caching is often appropriate for the idea of a request, but under-specified with respect to the \emph{constraints} that govern correctness. This is where lexical reasoning becomes useful: if the cache is keyed by (or at least mediated through) a lexical model, the matching problem is better posed.

\subsection{Hybrid Caching}
\label{sec:hybridcaching}

Hybrid approaches acknowledge the failure modes above and treat lexical and semantic evidence as complementary rather than competing signals. Exact matching provides high precision for duplicates; lexical retrieval (e.g., BM25 scoring~\cite{robertson2009bm25}) guards critical entities; dense retrieval recovers paraphrases and compositional variants. Prior work has shown that such mixtures can reduce latency and improve precision in real-world applications, particularly when a strict semantic-only cache yields too many near-miss collisions~\cite{Haqiq2025MinCache}.

While recent frameworks like Agentic Plan Caching~\cite{zhang2025costefficientservingllmagents} have begun to explore reuse beyond simple QA pairs, most hybrid caches are still architected as perimeter systems: they intercept the user prompt, retrieve a candidate complete response, and either return it or fall back to generation. \textbf{SemanticALLI} takes the hybrid logic and moves it inward. We use hybrid retrieval not only to decide whether a \emph{final} answer can be reused, but to target \emph{intermediate} artifacts at two checkpoints—AIR and VS—where repeated structure is empirically standard even when the user’s wording is not. That shift changes the unit of reuse from “the whole answer” to “the reusable parts,” which is precisely what monolithic caching cannot exploit.

\subsection{BI Systems}

Prior BI systems such as SiriusBI~\cite{jiang2025siriusbicomprehensivellmpoweredsolution} employ structured semantic intermediate representations to improve correctness and compositional reasoning. SemanticALLI is orthogonal to this line of work: rather than introducing a new analytic IR, our contribution is to treat such IRs as cacheable, persistent artifacts across user requests. The novelty lies not in intent decomposition itself, but in demonstrating that intermediate reasoning steps—particularly downstream synthesis—exhibit high reuse across otherwise distinct prompts, and that this reuse can be safely exploited via hybrid retrieval.

\section{Methods}
\label{sec:methods}

A monolithic cache assumes the application is a function from a natural-language prompt to a final response. That abstraction is convenient, and for lightweight chat it is often good enough. Dashboard generation is not lightweight. One “answer” is usually an assembly of parts: a resolved analytic specification, a set of charts, and code artifacts that are individually reusable across requests.

\textbf{SemanticALLI} therefore treats caching as an \emph{inference-time systems problem}. The key move is to identify stable checkpoints within the workflow—places where the system has already committed to an interpretable structure—and to cache the agentic system's intermediate steps, not just its terminal output.

\subsection{Two-Stage Decomposition and Cacheable Artifacts}

We decompose generation into two stages and cache the intermediate representations (IRs) for each stage. The decomposition is intentionally coarse: it is fine-grained enough to expose reuse, but not so fine-grained that cache management becomes the bottleneck.

\subsubsection{Stage I: Analytic Intent Resolution (AIR)}
AIR functions as a semantic normalization layer. Given a user prompt $q$ and context schema $S$, it produces a canonical \textbf{Analytic Intent Definition} $I$ that specifies \emph{what} to compute—metrics, dimensions, filters, temporal grain, chart primitives, and (when applicable) dashboard layout. Importantly, $I$ is designed to be stable under paraphrase and stylistic drift:
\begin{equation}
f_{\text{AIR}}(q, S) \rightarrow I .
\end{equation}

\textbf{SemanticALLI} caches this mapping at the level of intent. Practically, this means the cache stores $(q,S)\!\rightarrow\! I$. The goal is not to “freeze language,” but to absorb linguistic variance early so downstream synthesis can operate on a compact, structured object.

A small but consequential detail: AIR must remain entity-aware. In enterprise analytics, two requests can be semantically adjacent yet operationally incompatible. ``CPC'' and ``CPM'' may both live in the same neighborhood of an embedding space; the cache must not treat that neighborhood as equivalent. Our retrieval design reflects this constraint (Section~\ref{sec:hybrid_retrieval}).

AIR is not intended to be a universal analytic language; it is a task-specific, schema-grounded intermediate representation designed to stabilize downstream synthesis under paraphrase.

\subsubsection{Stage II: Visualization Synthesis (VS)}
VS takes the resolved intent $I$ and generates executable Visualization Directives C—e.g., chart code or dashboard composition instructions —through LLM agents:
\begin{equation}
f_{\text{VS}}(I) \rightarrow C .
\end{equation}

Here the lookup key is the \emph{structured intent} itself (or a deterministic serialization of it), not the original prompt. This difference is more than bookkeeping. It enables cross-query reuse even when prompts share no lexical overlap. Once two requests collapse to the same intent, VS can reuse prior synthesis without re-deriving the implementation layer.

\subsection{Hybrid Retrieval Engine}
\label{sec:hybrid_retrieval}

Both AIR and VS caching require retrieval that is fast, conservative with respect to entities, and robust to paraphrase. No single signal reliably meets all three requirements. We therefore use a hybrid engine that (i) prefers exact matches, (ii) falls back to semantic neighborhoods when appropriate, and (iii) reins in semantic drift with lexical evidence.

\subsubsection{Tier 0: Exact Hash Caching}
We first check for deterministic recurrence. For any input string (raw prompt for AIR; serialized intent for VS), we compute:
\[
k = \text{SHA-256}(\text{input}),
\]
and perform an $O(1)$ lookup. This tier is unapologetically strict; it exists to harvest obvious duplicates cheaply.

\subsubsection{Tier 1: Dense Semantic Indexing}
Exact matches are the exception, not the rule, so we also maintain a semantic index. Inputs are embedded in a dense 3072-dimensional vector space using OpenAI's embedding model  \texttt{text-embedding-3-large} \cite{openaiTextEmbedding3LargeDocs}, yielding $v \in \mathbb{R}^{d}$ with $d=3072$. For approximate nearest-neighbor search we employ HNSW graphs ~\cite{malkov2016hnsw} under cosine similarity. The similarity between a query vector $v_q$ and a candidate $v_c$ is:
\[
S_{\text{knn}}(q,c) = \frac{v_q \cdot v_c}{\|v_q\|\,\|v_c\|}.
\]

Dense retrieval gives recall, but it also creates the familiar “nearby but wrong” failure mode in high-cardinality business spaces. That is why we do not accept dense neighbors uncritically.

\subsubsection{Lexical Evidence and Reranking}
To preserve precision on critical entities (metrics and dimensions), we pair dense retrieval with lexical scoring (BM25) and reranking. In practice, we can require that candidates share the mandatory terms, and we can down-rank vector neighbors that omit them, even when their embeddings are close.
\paragraph{Example.}
The prompts ``Show \textbf{DDA Revenue} by channel'' and ``Show \textbf{GA4 Revenue}
by channel'' exhibit cosine similarity $\approx 0.96$, yet reference distinct
attribution models with incompatible metric definitions. Dense retrieval alone
would treat these as cache hits. Our lexical layer, scoring over schema tokens,
detects the metric mismatch (\texttt{dda\_revenue} vs.\ \texttt{ga4\_revenue})
and correctly rejects the candidate, ensuring that attribution-sensitive queries
are not conflated despite their structural similarity.

\subsubsection{Retrieval System}
For the complete retrieval system, the acceptance policy is stage-specific but conceptually simple: the system returns a cached artifact only when the best candidate clears a similarity threshold $\tau$ (for dense signals) and satisfies the lexical constraints induced by the domain (for schema and metric tokens). When these conditions are not met, we regenerate the artifact and optionally admit it to the cache under the configured admission policy.

% --- ALGORITHM BLOCK ---
\begin{algorithm}[ht]
\caption{Hybrid Retrieval with Reciprocal Rank Fusion (RRF)}
\label{alg:hybrid_retrieval_rrf}
\begin{algorithmic}[1]
\STATE \textbf{Input:} Query $q$, Cache Index $\mathcal{C}$, Threshold $\tau$, RRF Constant $k_{rrf}=60$
\STATE \textbf{Output:} Cached Artifact $a$ or \texttt{NULL}

\STATE \textit{// Tier 0: Exact Hash Lookup}
\STATE $h \leftarrow \text{SHA256}(q)$
\IF{$h \in \mathcal{C}_{\text{exact}}$}
    \STATE \textbf{return} $\mathcal{C}_{\text{exact}}[h]$
\ENDIF
\STATE \textit{// Tier 1: Parallel Retrieval}
\STATE $\mathcal{R}_{\text{dense}} \leftarrow \text{HNSW\_Search}(\mathcal{C}_{\text{dense}}, \text{Embed}(q), top\_k=10)$
\STATE $\mathcal{R}_{\text{lex}} \leftarrow \text{BM25\_Search}(\mathcal{C}_{\text{lex}}, q, top\_k=10)$

\STATE \textit{// Tier 2: RRF Fusion \& Reranking}
\STATE $\mathcal{U} \leftarrow  \mathcal{R}_{\text{lex}}$
\FOR{each candidate $c \in \mathcal{U}$}
    \STATE $r_d \leftarrow \text{Rank}(c, \mathcal{R}_{\text{dense}})$ \textit{// $\infty$ if not present}
    \STATE $r_l \leftarrow \text{Rank}(c, \mathcal{R}_{\text{lex}})$ \ \ \textit{// $\infty$ if not present}
    \STATE $score_{\text{rrf}}(c) \leftarrow \frac{1}{k_{rrf} + r_d} + \frac{1}{k_{rrf} + r_l}$
\ENDFOR

\STATE $\mathcal{S} \leftarrow \text{SortDescending}(\mathcal{U}, score_{\text{rrf}})$

\STATE \textbf{return} \texttt{NULL} \textit{// Cache Miss}
\end{algorithmic}
\end{algorithm}

\section{Results}
\label{sec:results}

We report two complementary views of reuse, evaluated on a dataset of 1,000 production prompts derived from a digital media marketing workload. The dataset spans a diverse ontology of media channels (e.g., paid search, social display, programmatic video) and KPIs. To ensure the evaluation reflects realistic reuse rather than artificial inflation, we constructed the split by temporally ordering user requests and partitioning them into a seed set ($N=500$) and a subsequent challenge set ($N=500$). This temporal split preserves natural distribution drift and ensures that the challenge set tests generalization to future queries rather than merely interpolating within a static batch.

One more thing to note, we do not ablate individual retrieval components in this study; our goal is to demonstrate the existence and magnitude of reuse at structured checkpoints rather than to optimize the retrieval stack itself. A systematic ablation of dense, lexical, and rank-fusion strategies is left to future work.

\subsection{Experimental Setup}
\label{sec:experimental-setup}

\textbf{Monolithic baseline (prompt$\rightarrow$output).}
We evaluate a full-output cache with exact and semantic matching. At a similarity threshold $\tau=0.90$, a prompt is considered a semantic hit if its nearest cached neighbor exceeds $\tau$; otherwise, the request is regenerated and then admitted to the cache.

\textbf{Pipeline-aware evaluation (AIR/VS).}
We evaluate \textbf{SemanticALLI} on a structured-intent challenge set of 500 prompts at $\tau=0.90$, and instrument cache behavior at the AIR and VS stages. For each stage we report (i) invocation counts, (ii) exact vs.\ semantic hit rates, (iii) the number of LLM-backed calls, and (iv) token usage.

\subsection{Monolithic \texorpdfstring{($f_{\text{AIR}}(q,S)\rightarrow I$)}{(AIR mapping)}}
\label{sec:monolithic-baseline}

Table~\ref{tab:monolithic-090} summarizes monolithic caching at $\tau=0.90$. The baseline is instructive but also fragile: when it hits, it is maximally efficient (an entire response is reused); when it misses, there is nothing to salvage.

\begin{table}[ht]
\caption{Monolithic prompt$\rightarrow$output cache behavior at $\tau=0.90$ (500 prompts).}
\label{tab:monolithic-090}
\vskip 0.15in
\begin{center}
\begin{small}
\begin{sc}
\begin{tabular}{lcc}
\toprule
\textbf{Metric} & \textbf{Count} & \textbf{\%} \\
\midrule
Exact hit     & 0  & 0.0 \\
Semantic hit  & 194 & 38.7 \\
\midrule
Total hit     & 194 & 38.7 \\
Miss          & 306 & 61.3 \\
\bottomrule
\end{tabular}
\end{sc}
\end{small}
\end{center}
\vskip -0.1in
\end{table}

Two observations follow. First, semantic matching contributes the majority of hits at this threshold. Second—and this becomes more salient in more complex settings—over 60\% of prompts still miss entirely, leaving monolithic caching with no mechanism for partial reuse.

\subsection{Stage-Level Reuse with SemanticALLI}
\label{sec:stage-level-results}

Table~\ref{tab:sirc-stage} reports cache behavior by stage for the structured-intent evaluation at $\tau=0.90$. AIR is intentionally the “hard” layer: it must map open-ended language onto a canonical analytic specification. VS, by contrast, operates on structured intent and therefore encounters repeated sub-structures (chart templates, standard encodings, and recurring layout primitives) even when the upstream language is novel.

\begin{table}[t]
\caption{Stage-level cache behavior for SemanticALLI at $\tau=0.90$ (500 prompts).}
\label{tab:sirc-stage}
\vskip 0.15in
\begin{center}
\begin{small}
\begin{sc}
% Use resizebox to force the table to fit within the column width
\resizebox{\columnwidth}{!}{
    \setlength{\tabcolsep}{2.5pt} % Slightly tighten spacing for better resize ratio
    \begin{tabular}{lcccc}
    \toprule
    \textbf{Stage} & \textbf{Invoc.} & \textbf{Exact hits} & \textbf{Semantic hits} & \textbf{LLM calls} \\
    \midrule
    AIR   & 500  & 0 (0.00\%)      & 194 (38.7\%)       & 306 \\
    VS    & 4{,}841 & 4{,}023 (83.10\%) & 0 (0.00\%)   & 818 \\
    \midrule
    Total & 5{,}341 & 4{,}023 (75.32\%) & 194 (38.7\%)   & 1{,}124 \\
    \bottomrule
    \end{tabular}
}
\end{sc}
\end{small}
\end{center}
\vskip -0.1in
\end{table}

The asymmetry is the point. At $\tau=0.90$, AIR reuse is rare (38.7\%), while VS reuse is common (83.10\%). Put differently: even when the system is forced to re-derive intent, it often does \emph{not} need to regenerate the downstream artifact from scratch.

Cache latency further sharpens the interpretation. VS exact hits are effectively free at inference-time scale (avg 2.94\,ms; p50 2.66\,ms; p95 5.29\,ms), so each VS hit corresponds to an avoided LLM call and a meaningful wall-clock reduction. The semantic hits observed in the AIR stage are slower (average of 440.39\,ms in our trace), which is still typically negligible compared to an LLM call.

\subsection{Token Accounting}
\label{sec:token-accounting}

Token usage is where stage-level caching becomes concrete. Table~\ref{tab:sirc-tokens} reports prompt, completion, and total token counts for the structured-intent evaluation. Note that VS averages are reported in two ways: (i) per invocation when cache hits are counted as zero tokens, and (ii) per LLM-backed VS invocation (i.e., true generations).

\begin{table}[ht]
\caption{Token usage by stage for SemanticALLI at $\tau=0.90$ (500 prompts).}
\label{tab:sirc-tokens}
\vskip 0.15in
\begin{center}
\begin{small}
\begin{sc}
% Use resizebox to force the table to fit within the column width
\resizebox{\columnwidth}{!}{
    \setlength{\tabcolsep}{2.5pt} % Slightly tighten spacing for better resize ratio
    \begin{tabular}{lrrrr}
    \toprule
    \textbf{Stage} & \textbf{Prompt tok.} & \textbf{Compl. tok.} & \textbf{Total tok.} & \textbf{Avg Tok./invoc.} \\
    \midrule
    AIR   & 1{,}016{,}795 & 946{,}058 & 1{,}962{,}853 & 3{,}925.71 \\
    VS    & 2{,}116{,}448 & 2{,}402{,}825 & 4{,}519{,}273 & 933.54 \\
    Total & 3{,}133{,}243 & 3{,}348{,}883 & 6{,}482{,}126 & 1{,}213.65 \\
    \bottomrule
    \end{tabular}
}
\end{sc}
\end{small}
\end{center}
\vskip -0.1in
\end{table}

Two details are easy to miss if one only looks at totals. First, AIR remains token-heavy in this configuration, which is consistent with its role as the semantic bottleneck. Second, VS caching dramatically reduces the effective per-invocation footprint: LLM-backed VS calls average 5{,}524.78 tokens in this trace, whereas the per-invocation average drops to 933.54 once cache hits are included. That gap is the operational value of intermediate reuse.

% --- Token accounting figure (ICML) ---
\begin{figure}[ht]
  \centering
  % Replace the filename below with your exported figure (PDF preferred for ICML).
  \includegraphics[width=\linewidth]{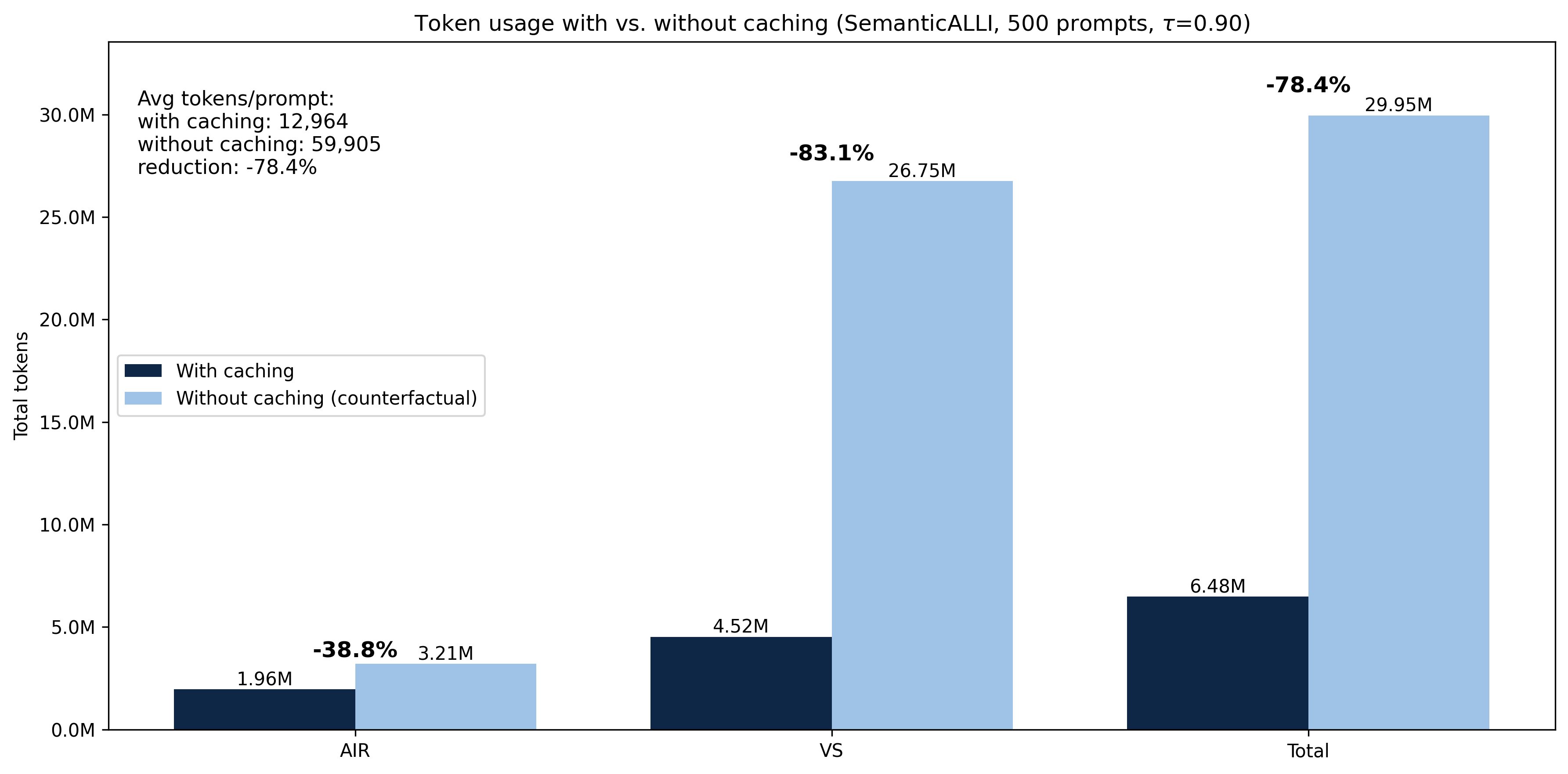}
  \caption{\textbf{Projected token usage with SemanticALLI caching vs.\ without caching (500 prompts, $\tau=0.90$).}
  ``Without caching'' is a counterfactual in which \emph{all} AIR prompts and VS invocations are LLM-backed at baseline per-call token costs (AIR: 6{,}414.55 tokens/prompt; VS: 5{,}524.78 tokens/invocation).
  ``With caching'' uses the observed SemanticALLI costs (AIR: 3{,}925.71 tokens/prompt; VS: 933.54 tokens/invocation).
  The projection uses the observed rate of VS invocations per user prompt ($4{,}841/500 \approx 9.68$), yielding average tokens per \emph{user prompt} of 59{,}906 without caching vs.\ 12{,}964 with caching (78.4\% reduction).
  Percent labels denote token reduction relative to the counterfactual.}
  \label{fig:tokens-500-projected}
\end{figure}

\begin{figure*}[ht]
  \centering
  % Replace the path/filename to wherever you store the exported PDF/PNG.
  \includegraphics[width=\textwidth]{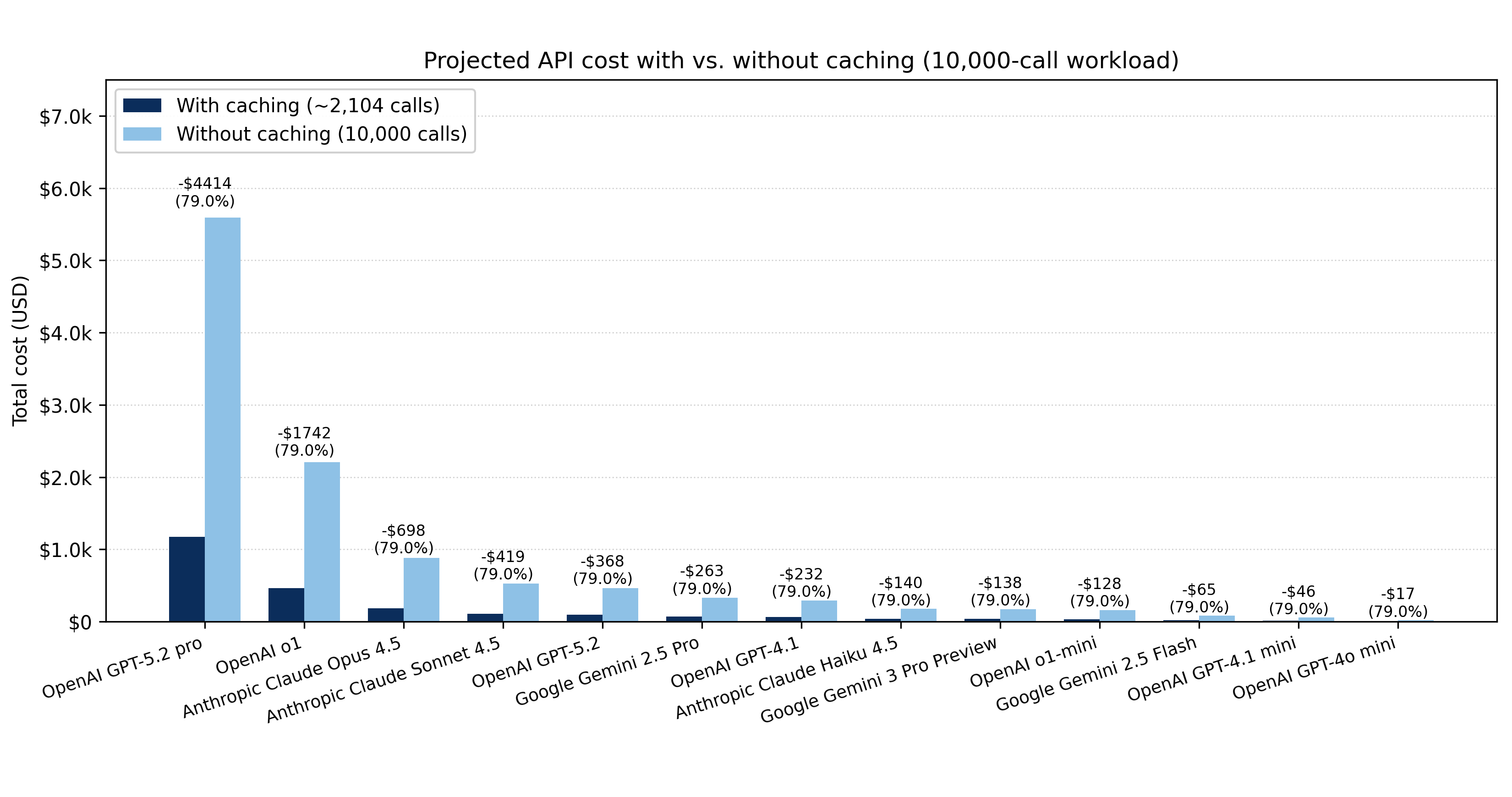}
  \caption{\textbf{Projected API cost with VS caching vs.\ without caching per 10{,}000 calls.}
  “With caching” assumes the measured call reduction at $\tau = 0.90$, retaining 21.04\% of LLM-backed calls ($\approx$2,104 of 10,000). Costs are computed using per-token list pricing (input/output) and the observed mean token footprint per call (2{,}788 input tokens; 2,979 output tokens).}
  \label{fig:cost_savings_10k_calls}
\end{figure*}

\begin{figure*}[!htbp]
  \centering
  % Replace with your exported PNG path
  \includegraphics[width=\textwidth]{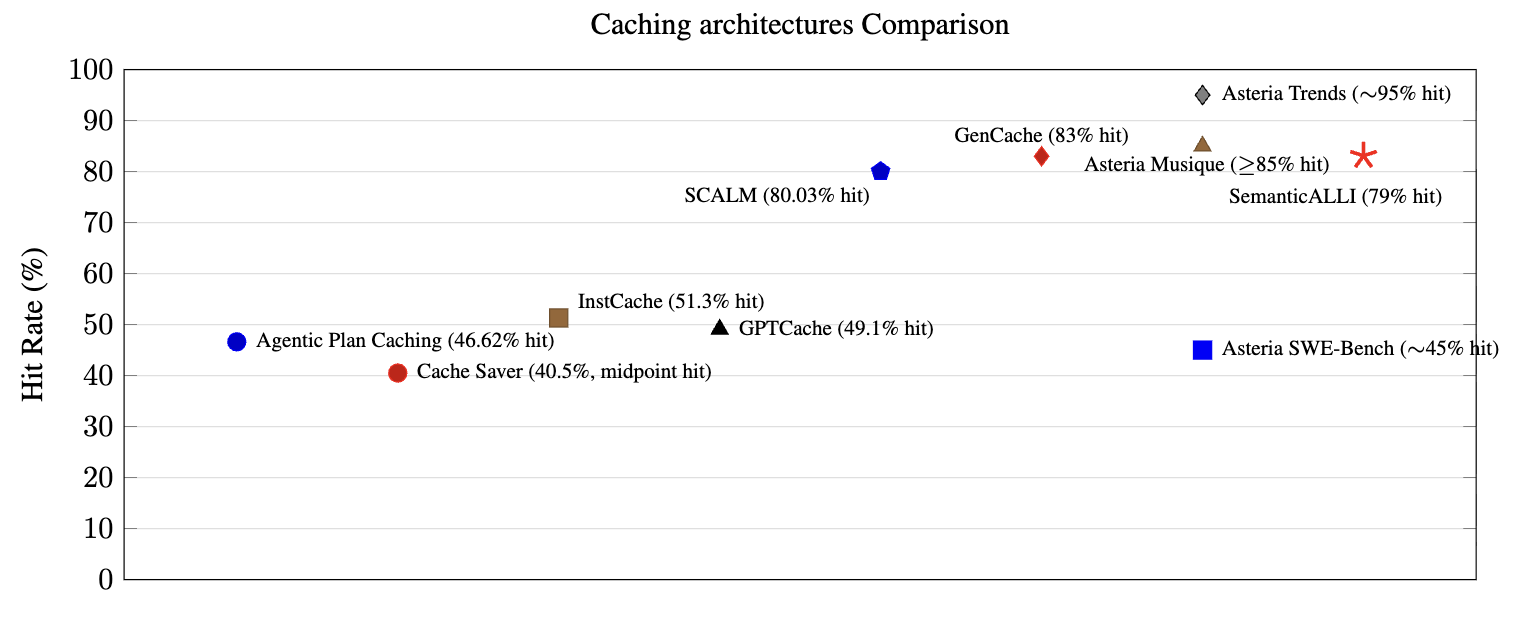}
  \caption{\textbf{Hit rate comparison for discussed caching architectures.}
  Cache Saver is shown at the midpoint of its reported 21--60\% range. Asteria reports workload-specific hit rates, shown as three points. SemanticALLI is marked with a star.}
  \label{fig:cache_scatter_with_point_labels}
\end{figure*}

\subsection{Threshold Sensitivity in End-to-End Latency}
\label{sec:threshold-sensitivity}

Finally, we examine how similarity thresholds interact with wall-clock behavior in practice. On the same set of 500 prompts evaluated at $\tau=0.90$ and $\tau=0.85$, lowering the threshold reduces end-to-end completion time substantially: mean runtime drops, for our ~20 agent system, from 57.3\,s to 31.5\,s and median runtime drops from 57.0\,s to 25.1\,s (p95: 87.5\,s to 61.5\,s).

This trade-off is expected. Lower thresholds are more permissive and thus increase reuse, but they also raise the risk of overly aggressive matching in entity-sensitive domains. The appropriate operating point depends on application tolerance for approximation versus strict correctness; in enterprise analytics, we generally treat entity-level fidelity as the binding constraint.

\section{Discussion}
\label{sec:discussion}

The goal of this work is not to maximize analytic accuracy, but to characterize and exploit reuse opportunities inside agentic pipelines. Accordingly, our evaluation focuses on reuse rates, avoided LLM calls, and token/latency reductions, rather than on end-to-end analytic accuracy. Therefore, in light of the previous statement, these results come from a closed test environment with our specific filters, so the expected results in other applications may vary due to the difference in similarity and re-usability of agentic components.

The monolithic baseline clarifies both the appeal and the brittleness of perimeter caching. At $\tau=0.90$, a prompt$\rightarrow$output cache delivers a 38.7\% hit rate (Table~\ref{tab:monolithic-090}); when it hits, it is maximally efficient because the system reuses the \emph{entire} output. The miss behavior is the problem. A 61.3\% miss rate means that for the majority of requests, monolithic caching contributes exactly nothing—no partial reuse, no amortization of repeated sub-decisions, just a cold-start generation path.

\textbf{SemanticALLI’s} stage-level results make the story of internal reuse more explicit. AIR reuse is scarce at the same threshold (Table~\ref{tab:sirc-stage}), which is not entirely surprising once one takes enterprise prompts seriously: intent resolution must reconcile synonyms, incomplete constraints, and schema-dependent assumptions, and two “similar” prompts can diverge on a single business-critical entity. In our business case, the AIR level is separated by media client and data source, making the low hit rate expected, as seen in Table~\ref{tab:sirc-stage}. For example, client A could use a name convention for metrics or dimensions that differs from client B, even though they may share the same intent.

The more interesting pattern sits downstream. VS exhibits heavy repetition—83.11\% exact-hit rate over 4{,}841 invocations (Table~\ref{tab:sirc-stage})—which implies that, in practice, users may explore a vast space of requests. At the same time, the system repeatedly instantiates a relatively small set of visualization and layout primitives. Thus, novelty at the language boundary does not imply novelty in the synthesis layer.

Note that at the VS level, caching involves chart types and columns, which are more likely to match exactly and, in turn, skip the hybrid level. At this level we typically use it for code generation or chart type selection, and we should not compromise on it; we set a high similarity threshold of $\tau = 0.95$. For example, a line chart should not include any semantically similar hits with intermediate artifacts of different chart types, as this would change the rendering of the user's intent entirely.

Token accounting reinforces this interpretation. Table~\ref{tab:sirc-tokens} shows that AIR remains expensive per invocation (3{,}925.71 tokens), reflecting its role as the semantic bottleneck. VS, however, is where caching converts directly into operational savings: once cache hits are counted as zero-token invocations, VS averages 933.54 tokens per invocation (Table~\ref{tab:sirc-tokens}), despite LLM-backed VS generations averaging 5{,}524.78 tokens in the same trace. That discrepancy is the practical difference between “the system must think from scratch” and “the system can retrieve a previously generated artifact in milliseconds.”

In Figure~\ref{fig:cache_scatter_with_point_labels}, we compare our pipeline-aware approach against a standard monolithic caching baseline (representative of strategies like GPTCache \cite{bang-2023-gptcache}). We observe a 69.25\% increase in hit rate, a gain achieved not by redesigning the underlying retrieval mechanics, but by rethinking where the cache resides within the inference architecture. By intercepting reasoning at stable intermediate checkpoints, we unlock reuse that boundary-level systems miss. We hypothesize that this "internal caching" paradigm is transferable and could yield similar efficiency gains for other multi-step agentic systems. We note that while these results highlight the structural advantage of our approach on a domain-specific dataset, further validation on diverse public benchmarks is necessary to generalize the findings.

Threshold choice complicates the picture in a way practitioners will recognize. In our paired latency analysis (Section~\ref{sec:threshold-sensitivity}), reducing $\tau$ from 0.90 to 0.85 materially improves end-to-end completion times on the same prompt subset. The temptation is to interpret this as a free win. It is not. Lower thresholds increase reuse, but they also increase the probability of reusing an artifact that is \emph{close} in semantic space while being wrong in analytic intent—an unacceptable failure mode when metric definitions, filters, and attribution logic are contractual rather than suggestive. This is precisely why hybrid retrieval (dense + lexical constraints) matters: it gives the system a mechanism to be permissive about phrasing while remaining conservative about entities.

Several limitations also emerge from these results. First, AIR’s low reuse at strict $\tau$ suggests that intent representations and similarity metrics still leave recall on the table; better canonicalization (especially for domain-specific metric aliases) and more schema-grounded representations may improve retrieval without relaxing correctness constraints. Second, the strong VS exact-hit behavior is partly a reflection of repeated templates, which is beneficial. Still, it raises an operational question about staleness when rendering libraries, schema mappings, or dashboard conventions evolve. Cache invalidation and versioning, therefore, become first-class concerns. Finally, our evaluation focuses on reuse under a fixed threshold; in deployment, thresholds may need to be adaptive (e.g., stricter when a query touches high-stakes metrics, looser for exploratory visual summaries).

Taken together, the results support a practical design claim: monolithic prompt$\rightarrow$output caching helps when the users repeat themselves. Still, pipeline-aware caching helps when the \emph{system} repeats itself—and in analytic agents, it often does.

\section{Conclusion}
\label{sec:conclusion}

As agentic AI systems are deployed more widely in industry, repeated user queries become the norm rather than the exception. Without reuse, this leads to redundant computation, increased token consumption, and longer response times. At the same time, product adoption depends on maintaining high user satisfaction even when complex multi-step workflows are involved.

\textbf{SemanticALLI} illustrates one such direction by storing and reusing structured intermediate representations throughout an analytic agent pipeline, as an example of what can be done more generally. By caching analytic intents, plans, and visualization artifacts, the system can avoid recomputing entire agentic flows when users rephrase or slightly modify their questions. Our evaluation shows that this approach can significantly reduce latency and token usage while preserving flexibility over natural language input.

Several limitations point to opportunities for future work. First, because our objective was to demonstrate a novel use-case for caching (intermediate reasoning) rather than a new retrieval algorithm, our results are specific to our production dataset. Future work should apply this pipeline-aware approach to standard public benchmarks to allow for direct comparison with other systems. Second, our reliance on client-specific caches simplifies isolation but prevents global pattern reuse. Given that KPIs often carry distinct meanings across clients, unlocking cross-tenant efficiency will require safe de-identification or controlled templates. Finally, high-cardinality data remains a challenge for intent resolution, potentially requiring dedicated embedding models, while the system itself can be extended via richer artifact types, learned admission policies, and sophisticated invalidation strategies. Ultimately, we hypothesize that this "internal caching" paradigm is transferable and could yield similar efficiency gains for a wide range of multi-step agentic systems. \footnote{SemanticALLI is a proprietary system developed and deployed internally at PMG; code, models, and production infrastructure are not publicly released.}

\section*{Impact Statement}
This paper presents work aimed at advancing the field of machine learning by improving the efficiency of multi-agent inference. By reducing token consumption and latency, this work may lower the carbon footprint of large-scale LLM deployments.

\section*{Acknowledgments}
The authors would like to thank  Abby Long, Anthony Pilleggi, Chris Davis, Crissi Cupak, Emily Fox, Kolby Morris, and Nathan Barling for their operational guidance and feedback throughout this project. We would also like to thank Avery Comer for being one of the best UI/UX folks we could have asked for throughout the project.

% Choose a style (plainnat, abbrvnat, unsrtnat, etc.)
\bibliographystyle{plainnat}

% Point to your .bib file (usually without the .bib extension)
\bibliography{references}

\end{document}